\documentclass[lettersize,journal]{IEEEtran}
\usepackage{amsmath}
\usepackage{amsfonts}
\usepackage{bbm}
\usepackage{algorithmic}
\usepackage{float}
\usepackage[linesnumbered,ruled]{algorithm2e}
\usepackage{array}
\usepackage{textcomp}
\usepackage{stfloats}
\usepackage{url}
\usepackage{verbatim}
\usepackage{graphicx}
\usepackage{subcaption}
\usepackage{cite}
\usepackage{colortbl}
\usepackage{pifont}
\usepackage{amssymb}
\usepackage{wrapfig}
\usepackage{physics}
\usepackage{xcolor}
\usepackage{xspace}
\usepackage{ragged2e} 
\usepackage{booktabs,makecell,multirow,tabularx}
\usepackage[colorlinks,linkcolor=blue,anchorcolor=blue,citecolor=blue]{hyperref}
\usepackage{cleveref}
\usepackage{bm}

\Crefname{equation}{Eq.}{Eqs.}
\Crefname{figure}{Fig.}{Figs.}
\Crefname{property}{Property}{Properties}
\begin{document}

\title{PointCloud-Text Matching: Benchmark Dataset and Baseline}

\author{Yanglin Feng, Yang Qin, Dezhong Peng, Hongyuan Zhu, Xi Peng and Peng Hu  
\thanks{This work was supported in part by the National Key R\&D Program of China under Grant 2024YFB4710604; in part by NSFC under Grant 62372315, 62472295, 62176171 and U21B2040; in part by Sichuan Science and Technology Planning Project under Grant 2024YFHZ0144, 2024YFHZ0089, 2024NSFTD0047 and 2024NSFTD0038; in part by System of Systems and Artificial Intelligence Laboratory pioneer fund grant; and in part by the Fundamental Research Funds for the Central Universities under Grant CJ202303 and CJ202403. \textit{(Corresponding author: Peng Hu.)}}
\thanks{Yanglin Feng, Yang Qin, Dezhong Peng, Xi Peng, and Peng Hu are with the College of Computer Science, Sichuan University, Chengdu 610065, China, Dezhong Peng is also with Sichuan National Innovation New Vision UHD Video Technology Co., Ltd, Chengdu 610095, China (e-mail: fcyzfyl@163.com; qinyang.gm@gmail.com;pengdz@scu.edu.cn; pengx.gm@gmail.com; penghu.ml@gmail.com).}
\thanks{Hongyuan Zhu is with the Institute for Infocomm Research (I$^2$R), A*STAR, Singapore (e-mail: hongyuanzhu.cn@gmail.com).}
}





\maketitle

\begin{abstract}
In this paper, we present and study a new instance-level retrieval task: PointCloud-Text Matching~(PTM), which aims to identify the exact cross-modal instance that matches a given point-cloud query or text query. PTM has potential applications in various scenarios, such as indoor/urban-canyon localization and scene retrieval. However, there is a lack of suitable and targeted datasets for PTM in practice. To address this issue, we present a new PTM benchmark dataset, namely SceneDepict-3D2T. We observe that the data poses significant challenges due to its inherent characteristics, such as the sparsity, noise, or disorder of point clouds and the ambiguity, vagueness, or incompleteness of texts, which render existing cross-modal matching methods ineffective for PTM. To overcome these challenges, we propose a PTM baseline, named \textbf{Ro}bust PointCloud-Text \textbf{Ma}tching method (RoMa). RoMa consists of two key modules: a Dual Attention Perception module (DAP) and a Robust Negative Contrastive Learning module (RNCL). Specifically, DAP leverages token-level and feature-level attention mechanisms to adaptively focus on useful local and global features, and aggregate them into common representations, thereby reducing the adverse impact of noise and ambiguity. To handle noisy correspondence, RNCL enhances robustness against mismatching by dividing negative pairs into clean and noisy subsets and assigning them forward and reverse optimization directions, respectively. We conduct extensive experiments on our benchmarks and demonstrate the superiority of our RoMa.

\end{abstract}


\section{Introduction}
\label{sec:intro}

\begin{figure}
    \centering
    \begin{minipage}[b]{0.307\linewidth}
        \centering
        \begin{subfigure}[b]{\linewidth}
            \includegraphics[width=1\linewidth]{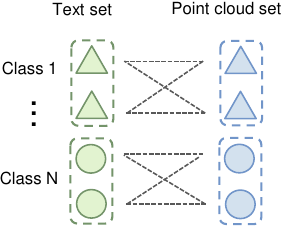}
            \caption{PTR.}\label{p2t_retrieval}
        \end{subfigure}
        \vfill
        \vspace{0.05cm}
        \begin{subfigure}[b]{\linewidth}
            \includegraphics[width=1\linewidth]{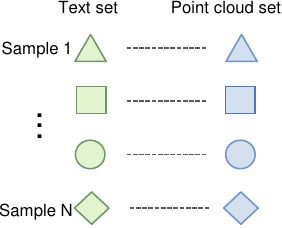}
            \caption{PTM.}\label{p2t_matching}
        \end{subfigure}
    \end{minipage}
    \hspace{0.05cm}
    \begin{minipage}[b]{0.562\linewidth}
        \centering
        \begin{subfigure}[b]{\linewidth}
            \includegraphics[width=0.99\linewidth]{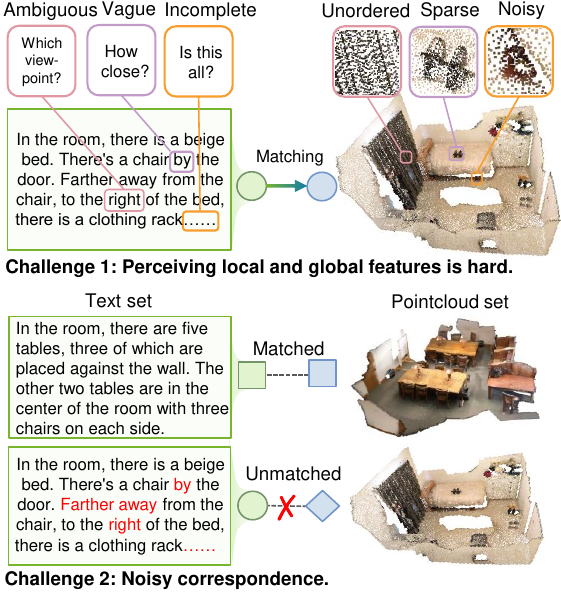}
            \caption{The challenges of PTM.}\label{p2t_md}
        \end{subfigure}
    \end{minipage}
    \caption{Overview for PointCloud-Text Matching (PTM). (a) and (b) show the schematic illustrations of class-level PointCloud-Text Retrieval (PTR), and instance-level PTM, respectively. (c) illustrates the challenges faced by PTM.}
    \label{task}
\end{figure}

\IEEEPARstart{P}{}oint clouds are a popular representation of the 3D geometry of a scene, with significant applications in computer vision, robotics, and augmented reality, such as autonomous driving~\cite{li2020deep, cui2021deep}, object detection~\cite{xu2021spg}, and localization~\cite{wang2023text}. However, as the volume of point-cloud data continues to grow rapidly, it is urgent to have techniques that enable users to effectively and accurately find the exact matching instance/scene from large-scale point-cloud scans, especially using natural language queries, named PointCloud-Text Matching~(PTM). The instance-level alignment is challenging and realistic as it reflects the need for precise and relevant information to build alignment between point clouds and texts in real-world applications, which has potential applications in indoor/urban-canyon localization, scene retrieval, and more.

Existing methods, however, struggle and lack pertinence to tackle PTM. On one hand, existing PointCloud-Text Retrieval (PTR) methods~\cite{huang2023joint,tang2023parts2words} only focus on establishing category-level correspondence between 3D point-cloud shapes and short annotation texts as shown in~\Cref{task}~\textcolor{blue}{(}\subref{p2t_retrieval}\textcolor{blue}{)}. In contrast, PTM requires exploiting the mutual information of cross-modal pairs, and achieves instance-level alignment between point-cloud scenes and detailed descriptions as shown in~\Cref{task} \textcolor{blue}{(}\subref{p2t_matching}\textcolor{blue}{)}. This indicates that PTM demands the ability to capture local features and instance discrimination, rendering the existing methods inapplicable. On the other hand, existing cross-modal matching works that can build instance-level cross-modal correspondence are only primarily oriented to text and 2D image modalities. According to the granularity of the established correspondence, these cross-modal matching works could be divided into two groups: coarse-grained and fine-grained matching methods. The former~\cite{faghri2017vse++, chen2021learning, qin2024cross} use global features to represent the whole image and the whole text, while the latter~\cite{liu2019focus,diao2021similarity,zhang2022negative} use local features to capture the fine-grained details of regions and words. Although these methods have achieved promising performance for image-text matching task, most of them ignore the specific properties and challenges in PTM.

\begin{figure}
    \flushleft
    \includegraphics[width=0.9\linewidth]{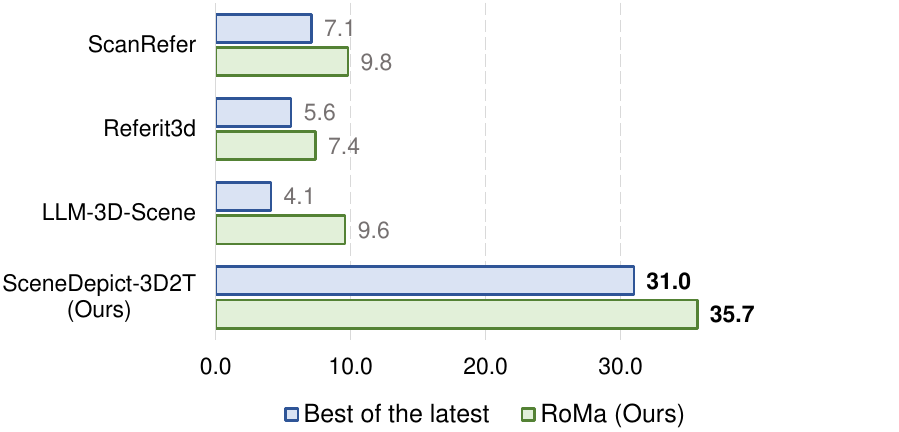}
    \caption{PTM performance (i.e., R@1) of the latest cross-modal matching methods (i.e., DIVE~\cite{kim2023improving}, CHAN~\cite{pan2023fine}, HREM~\cite{Fu_2023_CVPR}, and CRCL~\cite{qin2024cross}) and our RoMa on existing ScanRefer, Nr3d, 3D-LLM-Scene and proposed SceneDepict-3D2T dataset.}
    \label{data_baseline}
    \vspace{-0.2cm}
\end{figure}

To the best of our knowledge, the insufficient information provided by existing datasets makes them unsuitable for PTM. To be specific, descriptions in most datasets (e.g., ScanRefer~\cite{chen2020scanrefer}, Nr3d~\cite{achlioptas2020referit3d} primarily focus on portraying a single object for visual grounding and captioning, and a few other (e.g., LLM-3D-Scene~\cite{chen2024ll3da}) describing several objects in isolation within the corresponding scenes. 
These limited descriptions match precisely with the corresponding wide-field point clouds, as demonstrated by the dismal matching performance in existing datasets depicted in~\Cref{data_baseline}. Therefore, we constitute a new benchmark dataset for PTM, namely SceneDepict-3D2T. The dataset contains comprehensive descriptions covering entire 3D point-cloud scenes, so they evaluate baselines more reliably and reasonably for PTM, which can be observed in~\Cref{data_baseline}. We also provide a comprehensive evaluation protocol and several benchmark results for PTM on the datasets as shown in \Cref{compare_1}. From the results, we observe that point cloud-text data are more challenging than the common image-text data due to the sparsity, noise, or disorder of point clouds~\cite{liu2019deep}. More specifically, these properties make it difficult to capture and integrate local and global semantic features from both point clouds and texts and may also lead to mismatched cross-modal pairs, i.e., noisy correspondence~\cite{NEURIPS2021_f5e62af8,feng2023learning}, thus degrading the retrieval performance. The schematic illustration of the challenges is shown in~\Cref{task} \textcolor{blue}{(}\subref{p2t_md}\textcolor{blue}{)}. To be specific, the existing coarse-grained matching methods fail to extract discriminative global features from the unordered point clouds and vague texts, and the fine-grained matching methods that rely on well-detected object regions cannot be generalizable to point clouds. Moreover, most existing methods are based on well-annotated data and are susceptible to overfitting noisy correspondence, resulting in performance degradation. Therefore, there is a significant gap in applying existing methods to PTM.

To tackle the aforementioned challenges, we propose a PTM baseline, named \textbf{Ro}bust PointCloud-Text \textbf{Ma}tching method (RoMa), to learn from point clouds and texts as illustrated in \Cref{main}. RoMa consists of two modules: a Dual Attention Perception module (DAP) and a Robust Negative Contrastive Learning module (RNCL). DAP is proposed to adaptively capture and integrate the local and global informative features to alleviate the impact of noise and ambiguity in the data. More specifically, DAP conducts token-level and feature-level attention to adaptively weigh the patches and words to multigrainly aggregate the local and global discriminative features into common representations, thus embracing a comprehensive perception. In addition, our RNCL is presented to adaptively divide the negative pairs into clean and noisy subsets based on the similarity within pairs, and then assign them with forward and reverse optimization directions respectively. Different from traditional contrastive learning, our RNCL only leverages negative pairs rather than positive pairs to train the model since negatives are much less error-prone than positive pairs, leading to robustness against noisy correspondence. In brief, our RNCL could utilize and focus on more reliable pairs to enhance the robustness.

In summary, our main contributions are as follows:
\begin{itemize}
\renewcommand{\labelitemi}{$\bullet$}
    \item We investigate a new instance-level cross-modal retrieval task, namely PointCloud-Text Matching~(PTM), and propose a PTM benchmark dataset SceneDepict-3D2T and a robust baseline RoMa to learn from challenging multimodal data for PTM.
    \item We present a novel Dual Attention Perception module (DAP) that adaptively extracts and integrates the local and global features into common representations by using token-level and feature-level attention, thereby achieving a comprehensive perception of semantic features. 
    \item To handle noisy correspondence, we devise a Robust Negative Contrastive Learning module (RNCL) that adaptively identifies clean and noisy negative pairs, and assigns them correct optimization directions accordingly, thus preventing the model from overfitting noise.
    \item We conduct extensive comparison experiments on four pointcloud-text datasets. Our RoMa remarkably outperforms the existing methods without bells and whistles, demonstrating its superiority over existing methods.
\end{itemize}

\section{Related Work}
\label{sec:relatedwork}
\subsection{Cross-modal Retrieval}
Cross-modal retrieval aims to search the relevant results across different modalities for a given query, e.g., image-text matching~\cite{chen2021learning,diao2021similarity}, 2D-3D retrieval~\cite{li2024romo,feng2023rono}, and visible-infrared re-identification~\cite{wu2021discover}, etc. Most of these works learn a joint common embedding space by applying cross-modal constraints~\cite{oord2018representation,chechik2010large}, which aims to pull relevant cross-modal samples close while pushing the irrelevant ones apart. These methods could roughly be classified into two groups: 
1) Coarse-grained retrieval~\cite{faghri2017vse++,chen2021learning,Fu_2023_CVPR,chen2023inter,yang2024robust} typically learns shared subspaces to build connections between global-level representations, which align images and texts in a direct manner. 2) Fine-grained retrieval~\cite{lee2018stacked,diao2021similarity,pan2023fine} aims to model cross-modal associations between local feature representations, e.g., the visual-semantic associations between word tokens and image regions. Unlike them, in this paper, we delve into a less-touched and more challenging cross-modal scenario, i.e., Pointcloud-Text Matching (PTM), which argues for building cross-modal associations between 3D space and textual space.

\subsection{3D Vision and Language}
In contrast to image and language comprehension, 3D vision and language comprehension represent a relatively nascent frontier in research. Most existing works focus on using language to confine individual objects, e.g., distinguishing objects according to phrases~\cite{achlioptas2019shapeglot} or detecting individual objects~\cite{chen2019text2shape}. With the introduction of the ScanNet~\cite{dai2017scannet}, ScanRefer~\cite{chen2020scanrefer}, and Nr3d~\cite{achlioptas2020referit3d} datasets, more works have expanded their focus to encompass the 3D scenes. Some existing works~\cite{yuan2021instancerefer,he2021transrefer3d} have tried to locate objects within scenes based on linguistic descriptions, completing the task of 3D visual grounding. Recently, with the introduction of Scan2Cap~\cite{chen2021scan2cap}, some efforts~\cite{jin2023context} focus on providing descriptions for objects about their placement. This is also known as 3D dense captioning. Recently, a few preliminary solutions~\cite{huang2023joint,tang2023parts2words} for pointcloud-text retrieval have begun to emerge, which only establish common discrimination for coarse category-level alignment between point-cloud shapes and brief category label texts. However, these category-level methods could not be migrated to PTM. There are still scarce methods focusing on instance-level alignment and matching between wide-field point clouds and natural language texts, which requires excavating more detailed and discriminative connections within cross-modal pairs.

\section{PointCloud-Text Matching}
\label{sec:task}
In this paper, we introduce a novel 3D vision and language task, namely PointCloud-Text Matching (PTM). The input cross-modal data of the task involves the 3D point clouds and free-form description texts. 
The goal of PTM is to support bi-directional retrieval between point clouds and corresponding texts, achieving instance-level cross-modal alignment.

However, the task presents notable discrepancies and task-specific challenges, which can be summarized as follows:

\begin{itemize}
\renewcommand{\labelitemi}{$\bullet$}
    \item \textbf{Perceiving local and global semantic features is hard.} Since sensor sampling characteristics and biases, point clouds are commonly presented as a collection of sparse, noisy, and unordered points. Compared to 2D images, point clouds encapsulate a wealth of additional objects and spatial properties, which results in more incomplete and ambiguous description texts. Such complexity makes it harder for existing models to accurately perceive local and global semantic features from both modalities.
    \item \textbf{Noisy correspondence.} 
     Imperfect annotations are ubiquitous, even well-labeled datasets containing latent noisy labels, as shown by the existence of over 100,000 label issues in the ImageNet~\cite{deng2009imagenet} and 3\%-20\% annotation errors in the Conceptual Captions~\cite{sharma2018conceptual}. However, due to the limitations of human perception and description of 3D space, annotation workers are unintentionally inclined to use vague expressions (such as ‘near’, ‘close to’, etc.) to describe the details of the point clouds incorrectly, introducing more correspondence annotation (i.e., noisy correspondence). Such noise would lead to insufficient learning or noise overfitting for existing models.
\end{itemize}

\section{Benchmark Datasets: SceneDepict-3D2T}   
\label{datasets}

\vspace{-0.2cm}
\begin{figure}[H]
    \centering
    \includegraphics[width=0.9\linewidth]{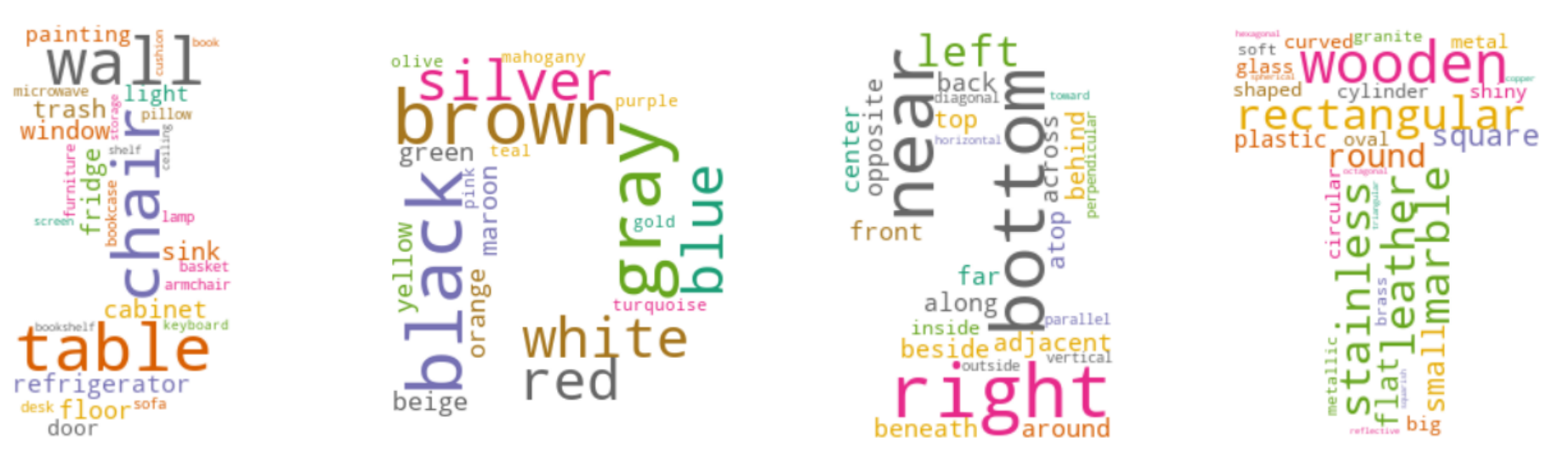}
    \caption{Word clouds of objects, colors, spatial information, and shape material in the descriptions of SceneDepict-3D2T.}
    \label{ciyun}
\end{figure}

To the best of our knowledge, existing multi-modal datasets of point clouds and texts can not apply directly to PointCloud-Text Matching (PTM). On the one hand, the descriptions in most of these datasets (e.g., ScanRefer~\cite{chen2020scanrefer}, Nr3d~\cite{achlioptas2020referit3d}, and ScanQA~\cite{azuma2022scanqa}) are confined to single objects of the entire point-cloud scenes. \Cref{bingtu} shows they only have average lengths of fewer than 15 words and each description encompasses fewer than 2 objects. However, one scene typically contains 10-30 objects~\cite{dai2017scannet}, with many similar objects present across different scenes. This indicates that these short and inadequately informative descriptions are prone to be ambiguous, lacking the discrimination to meet the requirements of PTM.
On the other hand, although several scene description datasets~\cite{chen2024ll3da} for scene understanding with the Large Language Model (LLM) have been recently proposed, they exhibit limited data volume and lack inter-object relationships necessary for a comprehensive description for each specific scene. These limitations hinder their application in PTM, as shown in \Cref{bingtu}. We conduct PTM experiments on these existing datasets, and the matching results in \Cref{data_baseline} show that the average performance of the latest cross-modal matching methods in terms of Recall at 1 is less than 10\%, confirming the above view.

To establish a reasonable evaluation of PTM with practical significance, we construct a diverse, detailed, and discriminative benchmark dataset with scene-level descriptions for comprehensive point-cloud scene understanding, namely SceneDepict-3D2T. 
In SceneDepict-3D2T, the point-cloud data is all based on the ScanNet~\cite{dai2017scannet} dataset, and the text data is derived from the ScanRefer~\cite{chen2020scanrefer}, Nr3d~\cite{achlioptas2020referit3d}, and ScanQA~\cite{azuma2022scanqa} description sets associated with ScanNet. In the following sections, we will elaborate on the collection and statistics of our proposed datasets.

\subsection{Data Collection}
\begin{figure}
    \centering
    \includegraphics[width=0.95\linewidth]{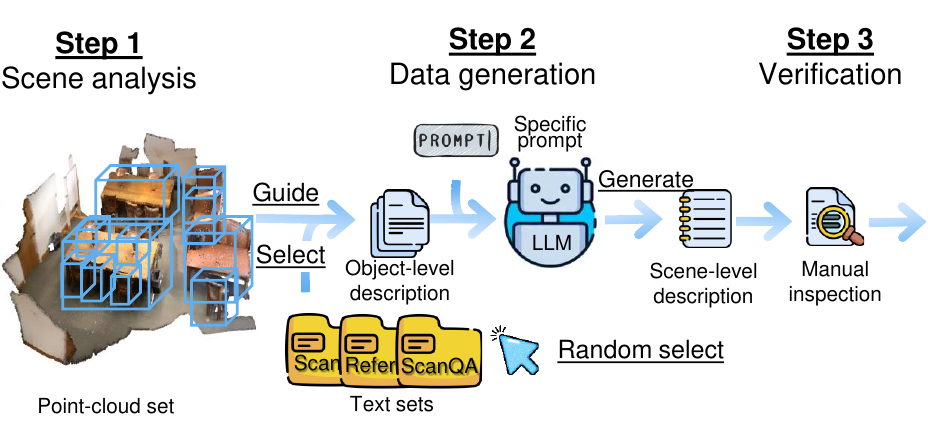}
    \caption{Pipeline of the one scene-level description collecting process in our SceneDepict-3D2T dataset.}
    \label{data_gen}
    \vspace{-0.1cm}
\end{figure}

We deploy a prompt-driven LLM paradigm to generate scene-level descriptions of point-cloud scene scans in ScanNet, leveraging three existing object-level description datasets (i.e., ScanRefer, Nr3d, and ScanQA). The description generation pipeline is divided into three stages, as illustrated in~\Cref{data_gen}.

1) \textit{Scene Analysis Stage}: We first divide each scene scan into multiple neighborhoods and identify discriminative objects based on their color, size, and more. We then randomly select $n$ descriptions of $n$ spatially related objects from different neighborhoods, creating object-level description collection. This stage arbitrarily introduces different object characteristics that guide the generated descriptions to encapsulate varied information and scene discrimination of point clouds.

2) \textit{Data Generation Stage}: In this stage, we randomly select one of the three object-level datasets and input the corresponding object-level description collection into the LLM using the tailored prompts to generate scene-level descriptions. The designed prompts align with the various linguistic characteristics of different datasets, ensuring the language style diversity and applicability of PTM across various scenarios. Specifically, for ScanRefer, the generated descriptions are objective and exhaustive, being suitable for scenarios with precise matching of the object placement throughout the scans. For Nr3d, the descriptions are concise and informative, being applicable for scenarios with matching of partial arrangement of objects within scans. For ScanQA, the descriptions detail object characteristics and relationships, which are suitable for scenarios with matching of key features of scans.

3) \textit{Verification Stage}: In this final stage, we manually assess each generated description for discriminative accuracy, grammatical correctness, and coherence, completing the scene-level description construction.

Repeating the above process ten times, we generate ten unique scene-level descriptions for each point cloud. By following this process for all point clouds, we create the SceneDepict-3D2T dataset. Note that more construction details and data examples are provided in the Appendix.

\subsection{Dataset Statistics}

\begin{figure}
    \centering
    \begin{subfigure}{0.33\linewidth}
        \includegraphics[width=1\linewidth]{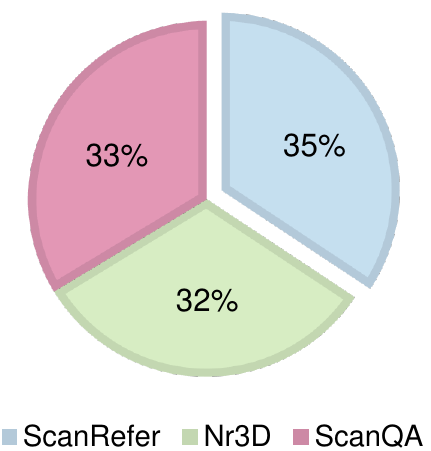}
        \caption{Corpus distribution}\label{xiaobingtu_1}
    \end{subfigure}
    \hspace{0.0cm}
    \begin{subfigure}{0.55\linewidth}
        \includegraphics[width=1\linewidth]{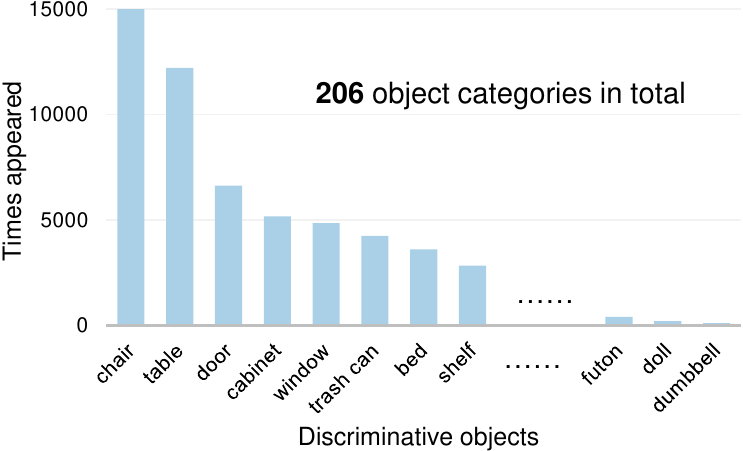}
        \caption{Scene object distribution}\label{xiaobingtu_2}
    \end{subfigure}
    \caption{Data distribution statistics. (a) shows the proportion of source corpus, and (b) shows the appeared times statistics of the discriminative objects in our SceneDepict-3D2T dataset.}
    \label{xiaobingtu}
    \vspace{-0.15cm}
\end{figure}

\begin{figure*}
    \centering
    \includegraphics[width=0.80\linewidth]{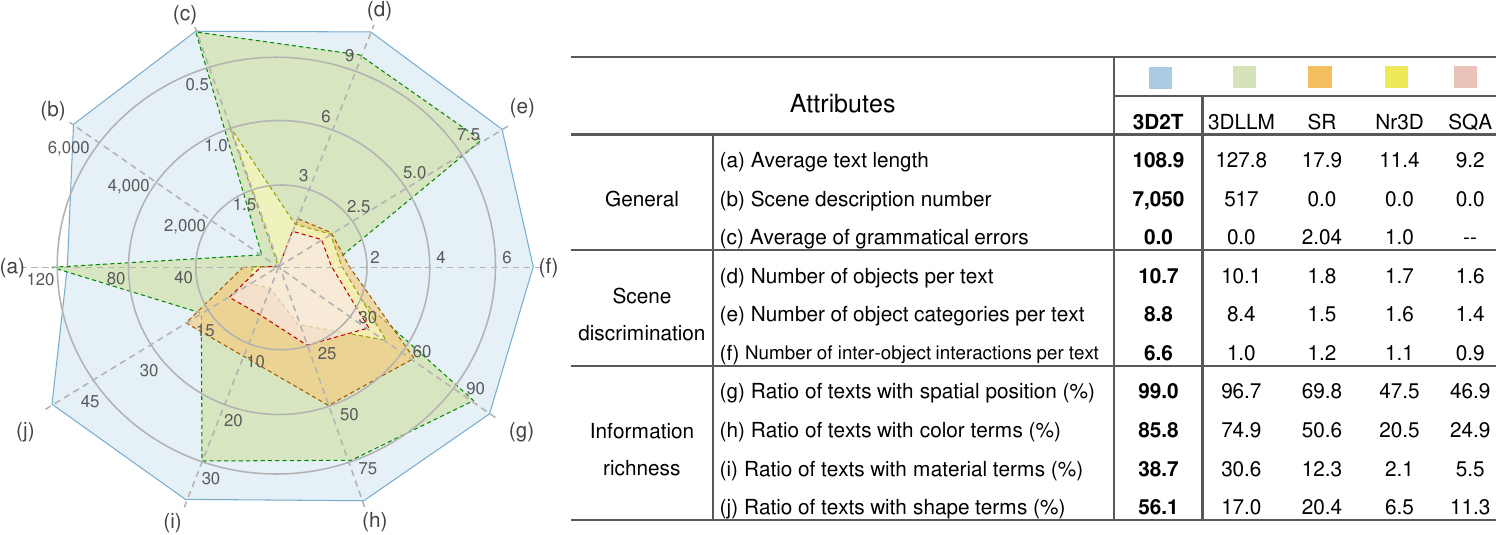}
     \caption{Statistics comparison among existing ScanRefer (SR)~\cite{chen2020scanrefer}, Nr3d~\cite{achlioptas2020referit3d}, ScanQA (SQA)~\cite{azuma2022scanqa}, 3D-LLM-Scene (3DLLM)~\cite{chen2024ll3da}, and proposed SceneDepict-3D2T (3D2T) dataset benchmarks.}
    \label{bingtu}
    \vspace{-0.15cm}
\end{figure*}

To provide a comprehensive overview of the proposed SceneDepict-3D2T dataset, we present data distribution statistics and compare it with those of existing datasets, as depicted in \Cref{xiaobingtu,bingtu}. 

More specifically, the corpus adopted for description generating is evenly sourced from the datasets with distinct linguistic styles, showcasing the comprehensiveness of the constructed descriptions. Additionally, our SceneDepict-3D2T offers a wealth of grammatical scene-level descriptions suitable for PTM training and validation.
On average, each description in SceneDepict-3D2T covers 10.7 objects and 8.8 object categories, which is over 5 times more than the object-level datasets (i.e., ScanRefer, Nr3d, and ScanQA). Furthermore, each description encompasses 6.6 inter-object interactions, which is 6.6 times higher than in the scene description dataset 3D-LLM-Scene. This demonstrates the sufficient scene coverage and discrimination of the descriptions in SceneDepict-3D2T. In addition, the descriptions in SceneDepict-3D2T are rich in color (85.8\%), material (38.7\%), shape terms (56.1\%), and spatial information (99.0\%), ensuring their informational depth. Consequently, benefiting from the discriminative and detailed descriptions in SceneDepict-3D2T, baselines can achieve 100\% to 400\% improvement in PTM performance compared to existing datasets, as shown in \Cref{data_baseline}, underscoring its practical significance for PTM.

Despite meticulous verification efforts to improve the grammatical accuracy and syntactic coherence of the datasets, it is unavoidable to introduce a considerable portion of noisy correspondence because of the inherent nature of unordered point-cloud scenes and vague free-form descriptions. To assess this, we randomly selected 100 descriptions from SceneDepict-3D2T and manually checked for vague expressions that might lead to noisy correspondence. Eventually, 13 descriptions were flagged for potential issues. Thus, noisy correspondence remains an unavoidable challenge in PTM, which could result in noise overfitting, leading to performance degradation.

\section{Robust Baseline: RoMa}
\subsection{Problem Formulation}
We first define the notations for a lucid presentation. Give a PTM dataset $\mathcal{D} = \{\mathcal{P}, \mathcal{T}\}$, where $\mathcal{P}=\{X^{p}_{i}\}^{N_p}_{i=1}$ and $\mathcal{T}=\{X^{t}_{j}\}^{N_t}_{j=1}$ are the point-cloud and text sets respectively, $N_p$ and $N_t$ are the size of $\mathcal{P}$ and $\mathcal{T}$, $X^{p}_{i}$ is the $i^{th}$ point-cloud sample and $X^{t}_{j}$ is the $j^{th}$ text sample. There is pairwise correspondence between $\mathcal{P}$ and $\mathcal{T}$, so $\mathcal{D}$ can also be written as $\mathcal{D}=\{(X^{p}_{i},X^{t}_{j}), y_{i,j}\}_{i,j}^{N_p, N_t}$, $y_{i,j}\in\{0, 1\}$ indicates whether $X^{p}_{i}$ and $X^{t}_{j}$ are matched (i.e., positive pair, $y_{i,j}=1$) or unmatched (i.e., negative pair, $y_{i,j}=0$). However, in practice, the unmatched pairs $(y_{i,j}=0)$ may be mislabeled as matched ones ($y_{i,j}=1$), \textit{a.k.a} noisy correspondence.

In the data encoding stage, we first employ modality-specific backbones (i.e., $f_p$ and $f_t$) to extract token-wise features for the patches of point clouds and words of descriptions, i.e., $Z^{p}_{i} = f_p(X^{p}_{i})\in \mathbb{R}^{p_n \times d_c}$ and $Z^{t}_{j}= f_t(X^{t}_{j})\in \mathbb{R}^{t_n \times d_c}$, respectively. $Z^{p}_{i}$ and $Z^{t}_{j}$ are the token-wise feature sets of $i^{th}$ point cloud and $j^{th}$ text, $p_n$ and $t_n$ are the number of tokens for each sample and $d_c$ is the dimensionality of the feature space.

In addition, to preserve spacial interactions across patches within point clouds, inspired by the sequence position representation~\cite{vaswani2017attention}, we attempt to encapsulate the 2D position information of patches into a position embedding. The embedding is then used for following comprehensive token-level attention calculation, as detailed below:
\begin{equation} \label{pos_emb}
   E_{i} = \{f(E^{x}_{i,1},E^{y}_{i,1}),\cdots,f(E^{x}_{i,p_n},E^{y}_{i,p_n})\},
\end{equation}
where $E_{i} \in \mathbb{R}^{p_n \times d_c}$ is the patch position embedding of the $i^{th}$ point-cloud sample, $f$ denotes the fusion method (e.g., summation, concatenation, etc.) for combining the 
patch position embeddings, and $E^{x}_{i,j}$ and $E^{y}_{i,j}$ are the patch position embeddings calculated from the horizontal and vertical coordinates of patch centroids in $j^{th}$ patch of $i^{th}$ point-cloud sample. These embeddings are computed as follows:
\begin{equation} \label{pos_ex}
    E^{x}_{i,j,\epsilon}= \sin (\frac{h_{i,j}\cdot p_n}{10000^{\frac{\epsilon}{d_c}}}),\quad E^{x}_{i,j,\omega}= \cos (\frac{h_{i,j}\cdot p_n}{10000^{\frac{\omega-1}{d_c}}}),
\end{equation}
\begin{equation} \label{pos_ey}
    E^{y}_{i,j,\epsilon}= \sin (\frac{v_{i,j}\cdot p_n}{10000^{\frac{\epsilon}{d_c}}}),\quad E^{y}_{i,j,\omega}= \cos (\frac{v_{i,j}\cdot p_n}{10000^{\frac{\omega-1}{d_c}}}),
\end{equation}
where $h_{i,j}$ and $v_{i,j}$ are the normalized horizontal and vertical coordinates of the corresponding patch centroids, and $\epsilon$ and $\omega$ refer to the even and odd dimensionality indices of $E^{x}_{i,j}$ and $E^{y}_{i,j}$, respectively.

\begin{figure*}
    \centering
    \includegraphics[width=0.82\linewidth]{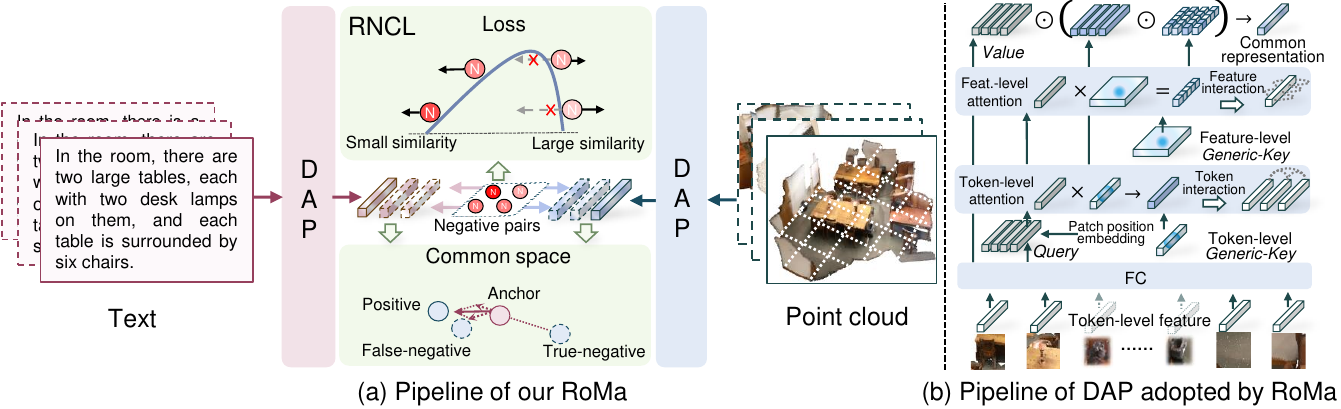}
    \vspace{-0.1cm}
     \caption{The illustration of our proposed method. (a) shows the pipeline of our RoMa, which involves two modules: Dual Attention Perception (DAP) and Robust Negative Contrastive Learning (RNCL). In DAP, comprehensive common representations could be extracted from both modalities and then matched into negative pairs. In RNCL, these negative pairs are adaptively optimized in both forward and reverse directions based on pairwise similarities, enhancing the robustness and discrimination of the common representations. (b) is the schematic illustration of DAP in point-cloud modality, which operates similarly for the text modality. \textit{Query} and \textit{Value} are obtained from features through a fully connected layer (FC), while \textit{Generic-Key} is general and learnable for the whole dataset. The \textit{Query} is combined with token-level and feature-level \textit{Generic-Key} to obtain dual attention. Following this, the features and attentions are aggregated into common representations.}
    \label{main}
    \vspace{-0.15cm}
\end{figure*}

To tackle the task-specific challenges mentioned earlier, a robust PTM method (RoMa) is proposed to learn cross-modal associations from point clouds and texts as shown in~\Cref{main}. The proposed method involves two modules: 1) Dual Attention Perception (DAP) is used to comprehensively perceive semantic features with dual attention at the dataset level, and 2) Robust Negative Contrastive Learning (RNCL) is exploited to handle noisy correspondence. In the following sections, we will elaborate on each component of RoMa.

\subsection{Dual Attention Perception}
 To tackle the challenge of capturing and integrating both local and global semantic features from point clouds and texts in PTM, we propose a novel dual attention mechanism. More specifically, similar to the definition in Self-Attention mechanism (SA)~\cite{vaswani2017attention}, we calculate the \textit{Queries} $Q^p_i \in \mathbb{R}^{p_n \times d_c}$, $Q^t_i \in \mathbb{R}^{t_n \times d_c}$, along with the \textit{Values} $V^p_{i} \in \mathbb{R}^{p_n \times d_c}$, $V^t_{j} \in \mathbb{R}^{t_n \times d_c}$ of two modalities from the $i^{th}$ point-cloud and $j^{th}$ text features $Z^{p}_{i}$ and $Z^{t}_{j}$, using the fully connected layers $g_p$ and $g_t$, respectively. However, unlike SA, our dual attention mechanism constructs learnable token-level and feature-level \textit{Generic-Keys}, which are shared across all samples in the dataset. Benefit from this, the \textit{Generic-Keys} could learn to model general patterns of tokens and features throughout the entire dataset. By integrating these \textit{Generic-Keys} with the sample-specific \textit{Queries}, our method achieves more comprehensive attention than SA, capturing interactions beyond token-wise relationships within individual samples, as depicted in~\Cref{main} \textcolor{blue}{(b)}.

To be more specific, to facilitate the adaptive exploration of local semantic features, we first construct token-level \textit{Generic-Keys} $\bar{K}^p \in \mathbb{R}^{d_c}$ and $\bar{K}^t \in \mathbb{R}^{d_c}$ upon the feature matrices of point-cloud and text modalities. We use them to model the common patterns of informative tokens (i.e., patches and words) within each modality. Similar to SA, we obtain token-level attention by measuring the token-wise similarity between \textit{Queries} and token-level \textit{Generic-Keys}, empowering the model to selectively focus on local key semantic units similar to the common patterns in the two modalities (e.g., foreground patches in the point clouds and keywords in the texts), which are written as: 
\begin{equation} \label{token_attn_vec}
   \bar{a}_{i}^{p} = \varphi((Q_{i}^{p} + E_{i})\bar{K}^{p\top}), \quad \bar{a}_{i}^{t} = \varphi(Q_{i}^{t} \bar{K}^{t\top}),
\end{equation}
where $\bar{a}^p_i \in \mathbb{R}^{p_n}$ and $\bar{a}^t_i \in \mathbb{R}^{t_n}$ are the token-level attention vectors, $\varphi$ is the Softmax operation along the token dimension (i.e., row-wise operation on the feature matrices). It is worth noting that we add patch position embedding in point-cloud modality to preserve the spatial interactions among patches. Based on this, we obtain the token-level attention $\bar{A}^p_i \in \mathbb{R}^{p_n \times d_c}$ and $\bar{A}^t_i \in \mathbb{R}^{t_n \times d_c}$ by stacking these attention vectors in the feature dimension.

In addition, we propose feature-level attention to capture feature semantics and enhanced cross-modal representations. Similar to token-level modeling, we introduce learnable feature-level \textit{Generic-Keys} $\hat{K}^p\in \mathbb{R}^{d_c \times d_c}$ and $\hat{K}^t\in \mathbb{R}^{d_c \times d_c}$ for two modalities, which aims to model the interaction patterns among $d_c$ features. We construct feature-level attention by combining \textit{Queries} and feature-level \textit{Generic-Keys} to grasp global discriminative features from the dimensional interrelationships in the feature space, such as distinctive object color, position, orientation, spatial relationships, etc., which is written as:
\begin{equation} \label{feature_attn}
    \hat{A}_{i}^{p} = \varphi(Q_{i}^{p} \hat{K}^{p\top}), \quad \hat{A}_{i}^{t} = \varphi(Q_{i}^{t} \hat{K}^{t\top}),
\end{equation}
where $\hat{A}^p_i \in \mathbb{R}^{p_n\times d_c}$ and $\hat{A}^t_i \in \mathbb{R}^{t_n\times d_c}$ are the feature-level attention.

Next, we aggregate the token-level and feature-level attention into dual attention, which can be written as:
\begin{equation} \label{attn}
    A^p_i=\bar{A}^p_i\odot \hat{A}^p_i,\quad A^t_i=\bar{A}^t_i\odot\hat{A}^t_i,
\end{equation}
where $A^p_i$ and $A^t_i$ are dual attention in point-cloud and text modalities, and the $\odot$ is the Hadamard product operator. Subsequently, we impose dual attention upon the \textit{Values}, aggregating them for integrated representations into common space, which are written as:
\begin{equation} \label{attn}
    \bm{p}_{i}=Norm(\frac{1}{p_n}\sum^{p_n}_{j} (A^p_{i,j} \odot V^p_{i,j})),
\end{equation}
\begin{equation} \label{attn}
    \bm{t}_{i}=Norm(\frac{1}{t_n}\sum^{t_n}_{j} (A^t_{i,j} \odot V^t_{i,j})),
\end{equation}
where $A^p_{i,j}$ and $A^t_{i,j}$ are the $j^{th}$ row of dual attention $A^p_i$ and $A^t_i$, $V^p_{i,j}$ and $V^t_{i,j}$ are the $j^{th}$ row of the \textit{Values} $V^p_{i}$ and $V^t_{i}$, and $Norm(\cdot)$ is the $l_2$-normalization function. The common representations $\bm{p}_{i}\in \mathbb{R}^{d_c}$ and $\bm{t}_{i}\in \mathbb{R}^{d_c}$ integrate local useful semantics and global discriminative semantics, promoting comprehensive feature perception in unordered point clouds and ambiguous texts.

\subsection{Robust Negative Contrastive Learning}
Inspired by \cite{10050111}, we leverage the complementary contrastive learning paradigm to learn with more reliable negative pairs instead of positive pairs, thereby mitigating the negative impact of mismatched pairs and achieving robust PTM against noisy correspondence. The loss for the cross-modal complementary learning paradigm is shown below:
\begin{equation} \label{L'}
\mathcal{L}' = \mathcal{L}^\prime_{p\to t} + \mathcal{L}^\prime_{t\to p},
\end{equation}
where
\begin{equation}
\mathcal{L}^\prime_{p\to t} = -\frac{1}{K} \sum^{K}_{i,j} (1-y_{i,j})\log{(1-S^{p\to t}_{i,j})},
\end{equation}
\begin{equation}
\mathcal{L}^\prime_{t\to p} = -\frac{1}{K} \sum^{K}_{i,j} (1-y_{i,j}) \log{(1-S^{t\to p}_{i,j})},
\end{equation}
and 
\begin{equation}
S^{p\to t}_{i,j} = \frac{\exp(\bm{p}_{i}^\top \bm{t}_{j}/\tau)}{\sum^{K}_{k} \exp(\bm{p}_{i}^\top \bm{t}_{k}/\tau)},
S^{t\to p}_{i,j} = \frac{\exp(\bm{t}_{i}^\top \bm{p}_{j}/\tau)}{\sum^{K}_{k} \exp(\bm{t}_{i}^\top \bm{p}_{k}/\tau)},
\end{equation}
where $\mathcal{L}^\prime_{p\to t}$\big/$\mathcal{L}^\prime_{t\to p}$ is the point-cloud-to-text/text-to-point-cloud complementary learning loss term, $S^{p\to t}_{i,j}$\big/$S^{t\to p}_{i,j}$ is the similarity between the $i$-th point-cloud/text sample and the $j$-th text/point-cloud sample, $K$ is the batch size, $\tau$ is the temperature parameter, and $1-y_{i,j}$ is the flag, making the loss only apply to negative pairs. Minimizing \Cref{L'} could reduce the similarity between the samples within negative pairs, introducing common discrimination without relying on positive pairs, which are more prone to containing some erroneous information. Because of this, the model could alleviate the impact of noisy correspondence.

\setlength\tabcolsep{3.7pt}
\begin{table*}[]
\caption{Performance comparison on the SceneDepict-3D2T dataset in terms of R@1, R@5, R@10 and the sum of them. The left side of the table shows the results of adopting Bi-GRU as the text backbone, and the right side shows the results of using BERT. The highest results are shown in \textbf{bold} and the second highest results are \underline{underlined}.}
\centering
\setlength{\tabcolsep}{3pt} 
\resizebox{0.81\textwidth}{!}{ 
\begin{tabular}{l|ccc|c|ccc|c||ccc|c|ccc|c}
\toprule
\multirow{3}{*}{Method}             & \multicolumn{8}{c||}{Bi-GRU}                                                                                                & \multicolumn{8}{c}{BERT}                                                                                \\
                                    & \multicolumn{4}{c}{Point cloud$\to$Text}                         & \multicolumn{4}{c||}{Text$\to$Point cloud}                             & \multicolumn{4}{c}{Point cloud$\to$Text}                 & \multicolumn{4}{c}{Text$\to$Point cloud}                 \\
\cmidrule(r){2-17}
                                    & R@1         & R@5         & R@10          & Sum            & R@1           & R@5           & R@10          & Sum          & R@1        & R@5        & R@10       & Sum         & R@1        & R@5        & R@10       & Sum       \\
\cmidrule(r){1-17}
VSE                                 & 9.9         & 35.5        & 47.5          & 92.9           & 11.1          & 35.5          & 48.6          & 95.2           & 23.4       & 48.2       & 57.2       & 128.8       & 16.2       & 48.7       & 62.0       & 126.9       \\
VSE++~\cite{faghri2017vse++}        & 14.9        & 36.2        & 49.6          & 100.7          & 11.8          & 36.3          & 50.1          & 98.2           & 21.3       & 43.3       & 56.7       & 121.3       & 17.2       & 47.3       & 62.6       & 127.1       \\
VSE$\infty$~\cite{chen2021learning} & 35.3        & 61.6        & 75.9          & 172.8          & 27.2          & 62.1          & 76.3          & 165.6          & 39.0       & 62.4       & 73.8       & 175.2       & 29.4       & 63.2       & 77.1       & 169.7       \\
SGR~\cite{diao2021similarity}       & 2.1         & 6.4         & 13.5          & 22.0           & 2.3           & 9.6           & 19.2          & 31.1           & 17.0       & 47.5       & 61.7       & 126.2       & 18.9       & 50.8       & 66.2       & 135.9       \\
SGR+NCR~\cite{NEURIPS2021_f5e62af8} & 5.8         & 17.1        & 37.3          & 60.2           & 7.0           & 23.0          & 43.8          & 73.8           & 29.8       & 58.2       & 65.2       & 153.2       & 21.5       & 56.4       & 71.8       & 149.7       \\
SAF~\cite{diao2021similarity}       & 9.9         & 34.0        & 48.9          & 92.8           & 12.8          & 37.5          & 53.2          & 103.5          & 21.3       & 53.2       & 67.4       & 141.9       & 20.1       & 56.5       & 66.1       & 142.7       \\
SAF+RCL~\cite{10050111}             & 17.6        & 44.0        & 60.3          & 121.9          & 17.1          & 44.3          & 59.9          & 121.3          & 31.9       & 57.4       & 70.9       & 160.2       & 22.1       & 55.6       & 71.4       & 149.1       \\
MV-VSE~\cite{li2022multi}           & 10.6        & 31.2        & 48.9          & 90.7           & 8.9           & 28.1          & 39.4          & 76.4           & 34.0       & 57.4       & 63.8       & 155.2       & 22.3       & 53.2       & 67.6       & 143.1       \\
NAAF~\cite{zhang2022negative}       & 17.7        & 41.1        & 52.5          & 111.3          & 10.9          & 31.6          & 45.0          & 87.5           & 17.0       & 44.0       & 59.6       & 120.6       & 13.8       & 44.3       & 63.1       & 121.2       \\
ESA~\cite{zhu2023esa}               & \underline{36.2}  & 66.0        & \underline{77.3}    & 179.5          & 25.8          & 62.6          & 75.5          & 163.9          & \underline{41.8} & \underline{70.2}       & \underline{81.4} & \underline{193.4} & \underline{32.0} & 68.6       & 79.9       & 180.5       \\
DIVE~\cite{kim2023improving}        & 28.4        & 66.7        & 77.1          & 175.2          & 22.7          & 61.3          & 67.8          & 145.8          & 37.6       & 67.4       & 76.6       & 181.6       & 28.4       & 65.3       & 78.0       & 171.7       \\
CHAN~\cite{pan2023fine}             & 28.8        & 66.6        & 68.3          & 188.7          & 22.1          & \underline{65.6}    & 64.4          & 120.5          & 40.1       & 62.7       & 76.3       & 179.1       & 26.6       & 58.2       & 73.4       & 158.2       \\
HREM~\cite{Fu_2023_CVPR}            & 34.0        & 59.6        & 69.5          & 163.1          & 27.7          & 64.8          & 78.0          & 170.5          & 39.0       & \textbf{70.9} & 81.3       & 191.2       & 31.5       & \underline{68.7} & \underline{81.2} & \underline{181.4} \\
CRCL~\cite{qin2024cross}            & 35.8        & \underline{67.8}  & 76.0          & \underline{179.6}    & \underline{28.8}    & 64.7          & \underline{77.1}    & \underline{170.6}    & \underline{41.8} & 64.5       & 80.1       & 186.4       & 30.5       & 64.8       & 77.4       & 172.7       \\
\cmidrule(r){1-17}
Ours                                & \textbf{42.0} & \textbf{73.0} & \textbf{84.4} & \textbf{199.4} & \textbf{29.3} & \textbf{68.9} & \textbf{82.1} & \textbf{180.3} &  \textbf{44.1}  & \textbf{70.9}  & \textbf{82.9}  & \textbf{195.9}  & \textbf{32.5}  & \textbf{71.0}  & \textbf{82.8} & \textbf{186.3} \\
\bottomrule
\end{tabular}
}
\label{compare_1}
\vspace{-0.2cm}
\end{table*}

However, due to the similarity in object categories within the point-cloud scenes and limited differences across some parts of scenes in PTM, samples within some negative pairs unavoidably exhibit certain degrees of semantic similarity. Blindly and monotonously amplifying the gap between two samples within negative pairs would lead to error accumulation, thus impacting the formation of robust discrimination. To address this issue, we propose the Robust Negative Contrastive loss, which could prevent the model from fitting these unreliable negative pairs or even revise the wrong optimization direction. This novel loss is non-monotonic and has a parameter-controlled inflection point. It assesses the reliability of negative pairs based on the similarity of the paired samples, dynamically and implicitly divides negative pairs into clean and noisy subsets based on their reliability by considering the inflection point as a threshold, and assigns clean subsets with forward optimization direction but provides noisy subsets with reverse optimization direction, which could be formulated as:
\begin{equation} \label{loss}
\mathcal{L} = \mathcal{L}_{p\to t}+ \mathcal{L}_{t\to p},
\end{equation}
where
\begin{equation}
\mathcal{L}_{p\to t} = -\frac{1}{K} \sum^{K}_{i,j}(1-y_{i,j})(1-S^{p\to t}_{i,j})^{\frac{1}{\alpha}} \log{(1-S^{p\to t}_{i,j})},
\end{equation}
\begin{equation}
\mathcal{L}_{t\to p} = -\frac{1}{K} \sum^{K}_{i,j} (1-y_{i,j})(1-S^{t\to p}_{i,j})^{\frac{1}{\alpha}} \log{(1-S^{t\to p}_{i,j})}.
\end{equation}
Note that $\mathcal{L}_{p\to t}$ and $\mathcal{L}_{t\to p}$ are the point-cloud-to-text and text-to-point-cloud loss terms of our Robust Negative Contrastive loss respectively, and $\alpha$ is the parameter that controls the inflection point. Take $\mathcal{L}_{p\to t}$ for example, its gradient calculation formula can be written as ${\partial \mathcal{L}_{p\to t}}/{\partial S^{p\to t}_{i,j}}=-\frac{1}{\alpha}(1 - S^{p\to t}_{i,j})^{\frac{1-\alpha}{\alpha}} [ \log(1 - S^{p\to t}_{i,j}) + \alpha]$. Consequently, we can infer that when $S^{p\to t}_{i,j}=1-e^{1-\alpha}$, the loss has a inflection point. As optimization progresses, clean negative pairs with low similarity (i.e., $S^{p\to t}_{i,j}<1-e^{1-\alpha}$) are still separated, while pairs with high similarity (i.e., $S^{p\to t}_{i,j}>1-e^{1-\alpha}$) are identified and brought closer, helping our RNCL filter out unreliable negative pairs adaptively. Compared to the existing loss~\cite{ross2017focal,zhang2018generalized}, our loss identifies and handles negative pairs, adaptively driving the reliable negative pairs apart in the common space, enhancing robustness against noisy correspondence in PTM.

\section{Experiments}
To thoroughly evaluate our RoMa for PTM, we conduct extensive experiments on the proposed SceneDepict-3D2T dataset and three other existing datasets.
\subsection{Experimental Settings}
In this work, our RoMa is implemented in PyTorch and carried out on one GeForce RTX 3090 GPU. In the experiments, we adopt the ScanNet~\cite{dai2017scannet} point-cloud set along with four description sets (i.e., SceneDepict-3D2T, ScanRefer~\cite{chen2020scanrefer}, Nr3d~\cite{achlioptas2020referit3d}, and 3D-LLM-Scene~\cite{chen2024ll3da}), obtaining corresponding four multi-modal datasets for PTM evaluation. Due to the space restriction, the details of implementation and introduction to the adopted datasets could be found in the Appendix.

\setlength\tabcolsep{3.6pt}
\begin{table*}[]
\caption{Performance comparison on the ScanRefer, Nr3d, and 3D-LLM-Scene datasets in terms of R@5, R@30 and the sum of them. The top of the table shows the results of adopting Bi-GRU as the text backbone, and the bottom shows the results of using BERT. The highest results are shown in \textbf{bold} and the second highest results are \underline{underlined}.}
\centering
\setlength{\tabcolsep}{3pt} 
\resizebox{0.82\textwidth}{!}{ 
\begin{tabular}{l|cc|c|cc|c|cc|c|cc|c|cc|c|cc|c}
\toprule
\multirow{3}{*}{Method}            & \multicolumn{6}{c|}{ScanRefer}                                                                   & \multicolumn{6}{c|}{Nr3d}                                                                 & \multicolumn{6}{c}{3D-LLM-Scene}                                                              \\
                                   & \multicolumn{3}{c}{Point$\to$Text}             & \multicolumn{3}{c|}{Text$\to$Point}             & \multicolumn{3}{c}{Point$\to$Text}            & \multicolumn{3}{c|}{Text$\to$Point}            & \multicolumn{3}{c}{Point$\to$Text}            & \multicolumn{3}{c}{Text$\to$Point}            \\
\cmidrule(r){2-19}
                                   & R@5           & R@30          & Sum            & R@5           & R@30          & Sum            & R@5           & R@30          & Sum           & R@5           & R@30          & Sum           & R@5           & R@30          & Sum           & R@5           & R@30          & Sum           \\
\cmidrule(r){1-19}
\multicolumn{19}{l}{\textbf{Bi-GRU:}} 
\\
VSE$\infty$~\cite{chen2021learning} & \underline{38.3}    & 77.3          & 115.6          & \underline{31.8}    & 73.0          & 104.8          & 24.8          & 55.3          & 80.1          & 21.3          & 54.3          & 75.6          & 6.6           & 37.7          & 44.3          & 9.8           & 50.8          & 60.6          \\
SAF+RCL~\cite{10050111}            & 29.8          & 68.1          & 97.9           & 28.0          & 71.8          & 99.8           & 19.9          & 52.5          & 72.4          & 19.8          & 56.7          & 76.5          & \underline{9.8}        & 42.6          & 52.4          & 9.4           & \textbf{57.5} & \underline{66.9}          \\
ESA~\cite{zhu2023esa}              & 34.8          & \underline{80.9}    & 115.7          & 31.6          & 71.5          & 103.1          & \underline{28.4}    & \underline{58.9}    & \underline{87.3}    & 21.9          & \underline{58.6}    & \underline{80.5}    & 6.6           & 42.6          & 49.2          & 7.9           & 54.1          & 62.0          \\
CHAN~\cite{pan2023fine}            & 31.9          & 69.5          & 101.4          & 25.0          & 67.7          & 92.7           & 19.9          & 47.5          & 67.4          & 15.9          & 43.5          & 59.4          & 6.9           & 39.3          & 46.2          & \underline{11.5}    & 51.6          & 63.1          \\
HREM~\cite{Fu_2023_CVPR}           & 33.3          & 71.6          & 104.9          & 27.3          & 68.9          & 96.2           & 19.9          & 53.9          & 73.8          & 19.1          & 53.1          & 72.2          & 8.2    & 34.4         & 42.6    & 10.7    & 46.7          & 57.1    \\
CRCL~\cite{qin2024cross}           & 35.5          & \underline{80.9}    & \underline{116.4}    & 31.3          & \underline{75.4}    & \underline{106.7}    & 20.6          & 57.1          & 77.7          & \underline{22.6}    & 53.6          & 76.2          & 6.6           & \underline{45.9}    & \underline{52.5}       & 9.0           & \underline{57.4}    & 66.4          \\
\cmidrule(r){1-19}
Ours                               & \textbf{41.8} & \textbf{82.3} & \textbf{124.1} & \textbf{33.1} & \textbf{76.0} & \textbf{109.1} & \textbf{31.9} & \textbf{61.7} & \textbf{93.6} & \textbf{26.0} & \textbf{61.3} & \textbf{87.3} & \textbf{15.7} & \textbf{47.5} & \textbf{63.2} & \textbf{16.2} & 55.3          & \textbf{71.5} \\
\cmidrule(r){1-19}
\cmidrule(r){1-19}
\multicolumn{19}{l}{\textbf{BERT:}}
\\
VSE$\infty$~\cite{chen2021learning} & \underline{44.6}    & 82.3          & \underline{126.9}    & 33.3          & 76.5          & 109.8          & \underline{25.5}    & 57.0          & 82.5          & 21.5          & 53.4          & 74.9          & 8.2        & 31.1       & 39.3       & \underline{9.8}          & 52.5        & 62.3  \\
SAF+RCL~\cite{10050111}            & 34.0          & 83.0          & 117.0          & 32.7          & \underline{78.9}    & 111.6          & 18.4          & 56.0          & 74.4          & 18.0          & \underline{58.7}    & 76.7          & 8.2        & 44.5       & 52.7       & 8.2          & 54.9        & 63.1  \\
ESA~\cite{zhu2023esa}              & 43.4          & 83.3          & 126.7          & \textbf{34.3} & 76.7          & 111.0          & 24.1          & 48.9          & 73.0          & 20.3          & 53.1          & 73.4          & 6.6        & \underline{46.5}       & 53.1       & \underline{9.8}     & \textbf{58.6}  & \underline{70.5}  \\
CHAN~\cite{pan2023fine}            & 36.9          & 79.5          & 116.4          & 32.3          & 70.8          & 103.1          & 22.6          & 50.4          & 73.0          & 15.0          & 46.1          & 61.1          & \underline{13.1} & 45.8 & \underline{58.9} & 8.2          & 47.8        & 56.0  \\
HREM~\cite{Fu_2023_CVPR}           & 40.4          & \underline{83.7}    & 124.1          & \underline{34.0}    & 77.9          & \underline{111.9}    & 24.7          & 52.5          & 77.2          & 18.1          & 54.0          & 72.1          & 9.8      & 36.1       & 45.9      & 9.0   & 54.1  & 63.1  \\
CRCL~\cite{qin2024cross}           & 42.6          & 83.0          & 125.6          & 32.4          & 77.0          & 109.4          & \underline{25.5}    & \underline{58.2}    & \underline{83.7}    & \underline{22.5}    & 57.7          & \underline{80.2}    & 7.6        & 39.3       & 46.9       & 9.0          & 55.7        & 64.7  \\
\cmidrule(r){1-19}
Ours                               & \textbf{49.6} & \textbf{88.3} & \textbf{137.9} & 32.8          & \textbf{84.3} & \textbf{117.1} & \textbf{31.9} & \textbf{58.9} & \textbf{90.8} & \textbf{27.9} & \textbf{65.6} & \textbf{93.5} &  \textbf{16.2}  & \textbf{47.5}  &  \textbf{63.7} &   \textbf{16.1} &  \underline{58.2}  &   \textbf{74.3}\\
\bottomrule
\end{tabular}
}
\label{compare_2}
\vspace{-0.2cm}
\end{table*}

\setlength\tabcolsep{2.pt}
\begin{table}{}
\begin{center}
\caption{Ablation studies for our RoMa framework and DAP module adopted by our RoMa on the SceneDepict-3D2T dataset. $\checkmark$ stands for use. \textit{w/o} stands for component removal.}
\setlength{\tabcolsep}{2pt} 
\resizebox{.43\textwidth}{!}{
    \begin{tabular}{ccc|ccc|cc|cc|c}
\toprule 
\multicolumn{3}{c|}{Feat. Extraction} & \multicolumn{3}{c|}{Loss}                   & \multicolumn{2}{c|}{Point$\to$Text}                    & \multicolumn{2}{c|}{Text$\to$Point}  &                  \\
\cmidrule(r){1-11} 
GPO           & ESA           & DAP          & \ \ $ \mathcal{L}_c $ \ \        & \ \ $ \mathcal{L}' $\ \     &\ \ $\mathcal{L} $ \ \    & R@1           & R@5              & R@1           & R@5          & Sum          \\
\cmidrule(r){1-11} 
 &        &   &  &              & $\checkmark$ & 1.4       & 9.9         & 1.1     & 7.6    &  20.0       
   \\
  &               & $\checkmark$  &  &              &              & 9.7      & 36.5        & 8.2    & 37.4   &  91.8      \\
\cmidrule(r){1-11} 
$\checkmark$  &               &               & $\checkmark$ &              &              & 35.3          & 61.6                    & 27.2          & 62.1          & 186.4          \\
$\checkmark$  &               &               &              & $\checkmark$ &              & 36.8          & 64.5                    & 27.4          & 64.1          & 192.8          \\
$\checkmark$  &               &               &              &              & $\checkmark$ & 37.2          & 66.0                    & 27.4          & 65.9          &  196.5         \\
              & $\checkmark$  &               & $\checkmark$ &              &              & 36.2          & 66.0                    & 25.8          & 62.6          & 190.6          \\
              & $\checkmark$  &               &              & $\checkmark$ &              & 39.7          & 67.4                    & 26.7          & 66.0          & 199.8          \\
              & $\checkmark$  &               &              &              & $\checkmark$ & 40.8          & 68.1                    & 28.1          & 67.5          & 204.5          \\
              &               & $\checkmark$  & $\checkmark$ &              &              & 37.4          & 59.6                    & 27.4          & 63.2          & 187.6          \\
              &               & $\checkmark$  &              & $\checkmark$ &              & 43.3          & 65.2                    & 27.4          & 66.7          & 202.6          \\                      
              \cmidrule(r){1-11} 
              &               & \textit{w/o} $\bar{A}$&              &  & $\checkmark$ & 41.6          & 72.7                    & 28.1          & 64.8          & 208.2          \\
              &               & \textit{w/o} $\hat{A}$&              &  & $\checkmark$ & 29.1          & 61.7                    & 21.6          & 57.7          & 170.1          \\
              &               & \textit{w/o} $E$ &              &  & $\checkmark$ & 37.6          & 72.0                    & 28.5          & 68.0          & 206.1          \\
              \cmidrule(r){1-11}
              &               & $\checkmark$  &              &              & $\checkmark$ & \textbf{42.0} & \textbf{73.0} & \textbf{29.3} & \textbf{68.9} & \textbf{213.2} \\
              
\bottomrule
\end{tabular}
}
    \label{ablation}
\end{center}
\vspace{-0.3cm}
\end{table}

In the experiments, we compared our RoMa with 14 state-of-the-art cross-modal matching methods, including VSE, VSE$++$~\cite{faghri2017vse++}, VSE$\infty$~\cite{chen2021learning}, SGR~\cite{diao2021similarity}, NCR-SGR~\cite{NEURIPS2021_f5e62af8}, SAF~\cite{diao2021similarity}, RCL-SAF~\cite{10050111}, MV-VSE~\cite{li2022multi}, NAAF~\cite{zhang2022negative}, DIVE~\cite{kim2023improving}, CHAN~\cite{pan2023fine}, ESA~\cite{zhu2023esa}, HREM~\cite{Fu_2023_CVPR}, and CRCL~\cite{qin2024cross}. In the implementations and evaluations of all the methods, we adhere to the following settings. For point-cloud data processing, we adopt the widely used DGCNN~\cite{wang2019dynamic} to obtain patch-level features. For text data processing, without loss of generality, we follow the text processing strategy used in cross-modal matching~\cite{chen2021learning,zhu2023esa} and employ both Bi-GRU~\cite{cho2014learning} and BERT~\cite{devlin2018bert} to acquire word-level features, respectively. We follow~\cite{lee2018stacked, NEURIPS2021_f5e62af8} to compute Recall at K (R@K) as the measurement of performance. In the experiments on SceneDepict-3D2T, we report R@1, R@5, R@10, and their sum to evaluate the performance of the methods. Due to the text discrimination limitations of other existing datasets (i.e., ScanRefer, Nr3d, and LLM-3D-Scene), we report R@5, R@30, and their sum to evaluate the methods more reasonably.

\subsection{Comparison with the State-of-the-Arts}
We conduct extensive PTM experiments on four datasets to evaluate the performance of our RoMa and the baselines. The experimental results are reported in~\Cref{compare_1,compare_2}. These results could yield the following observations: 1) General cross-modal matching methods exhibit inadequate performance. This substantiates the presence of distinct and more formidable challenges in PTM, indicating the difficulty of effectively applying these methods in PTM. 2) Some fine-grained methods (e.g., SGR and SAF) suffered from severe performance issues in PTM. By combining these methods with robust modules, such as NCR-SGR and RCL-SAF, the performance could be remarkably improved. These results indicate that there is a large amount of noisy correspondence in PTM, which leads to the performance degradation of the non-robust methods. 
\begin{figure}
    \centering
    \begin{subfigure}{0.40\linewidth}
        \includegraphics[width=1\linewidth]{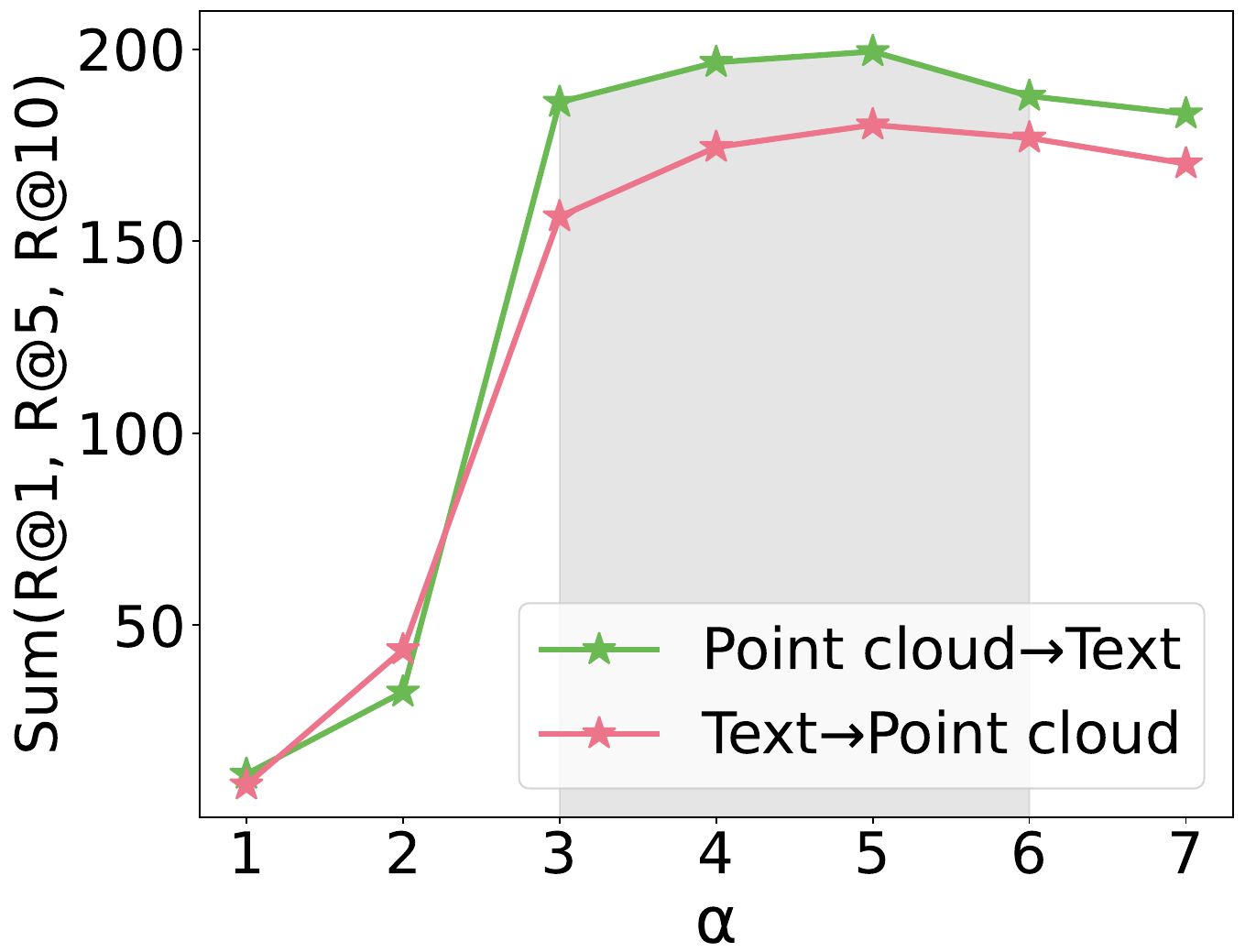}
        \caption{Impact of different $\alpha$.}\label{par_1}
    \end{subfigure}
    \hspace{0.3cm}
    \begin{subfigure}{0.40\linewidth}
        \includegraphics[width=1\linewidth]{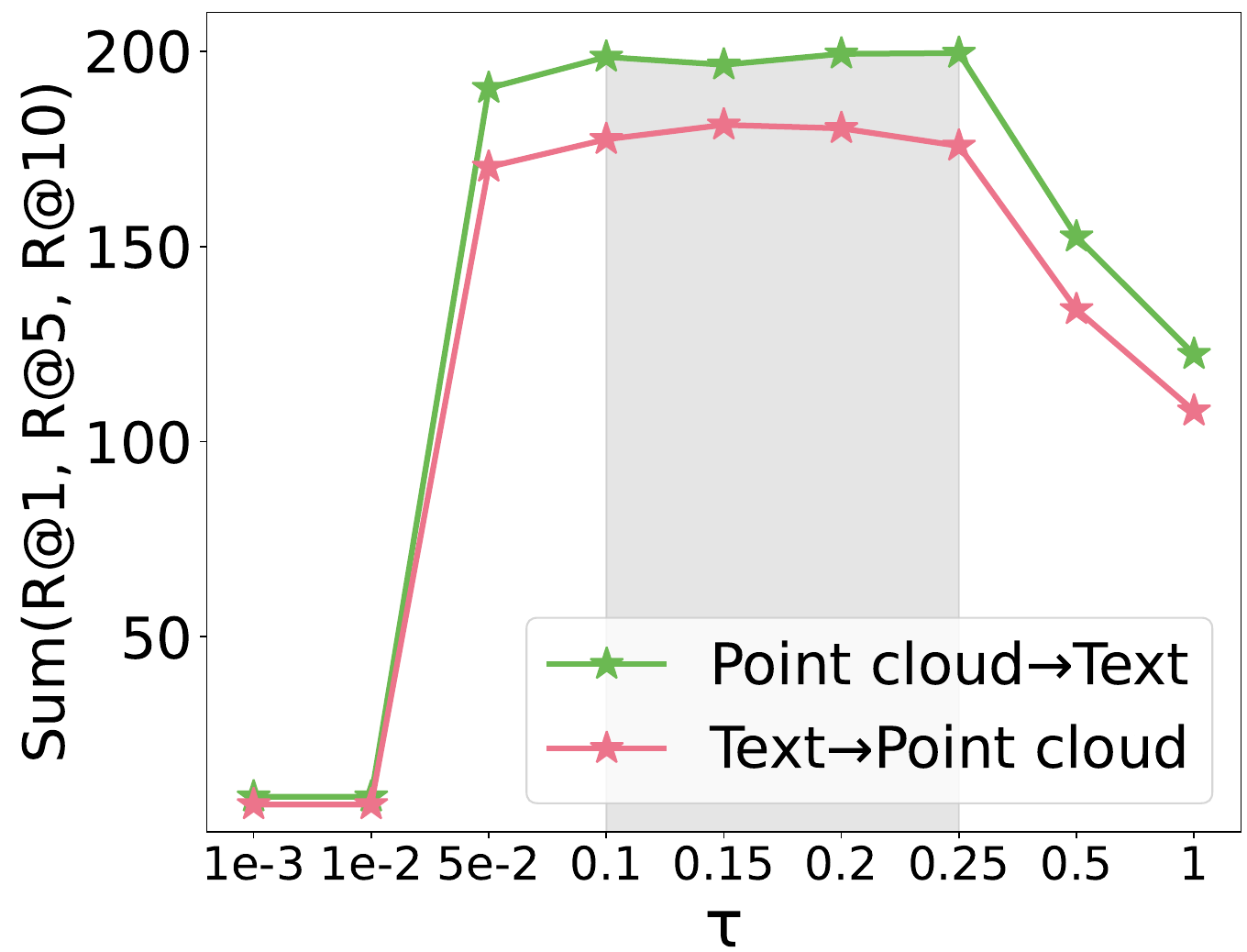}
        \caption{Impact of different $\tau$.}\label{par_2}
    \end{subfigure}
    \caption{Performance of RoMa in terms of the sum of R@1, R@5, R@10 versus different values of $\alpha$ and $\tau$ on SceneDepict-3D2T. The \textcolor[HTML]{9B9B9B}{\textbf{gray box}} is the optimal choice range.}
    \label{par}
    \vspace{-0.3cm}
\end{figure}
3) Our RoMa achieves remarkably better results than the existing cross-modal matching methods (e.g., CRCL, HERM, etc.), demonstrating its superior effectiveness by conquering the two challenges in PTM. 4) In existing datasets, the relevant results matched by most methods rank only outside the top 5, proving the scene-specific discrimination of these datasets is limited. 5) The performance on SceneDepict-3D2T is relatively low, compared to the existing Image-Text datasets, where the state-of-the-art performance usually exceeds 80~\cite{zhu2023esa,Fu_2023_CVPR}, in terms of R@1. This indicates that the PTM task still faces difficulties in handling unordered point clouds, vague texts, and noisy correspondence, and calls for more advanced solutions.

\subsection{Ablation Study}
In this section, we conduct an ablation study to investigate the contribution of each proposed component to PTM. Firstly, we replace the DAP module with the GPO~\cite{faghri2017vse++} and ESA~\cite{zhu2023esa} feature extraction modules, and the Robust Negative Contrastive loss (i.e., $\mathcal{L}$) adopted by RNCL with the vanilla loss adopted by complementary contrastive learning paradigm (i.e., $\mathcal{L}'$)~\cite{10050111} and Contrastive loss (i.e., $\mathcal{L}_{c}$). In addition, we alternately replace patch position embedding $E$, token-level attention $\bar{A}$, and feature-level attention $\hat{A}$ to fairly verify their effectiveness under the premise of eliminating the influence of the number of learnable parameters. All the comparisons are conducted on SceneDepict-3D2T with the same experimental settings. The results are presented in \Cref{ablation}. From the table, we could draw the following observation: 1) RoMa without any component will drop matching performance, which indicates that each component contributes to our method. 2) The performances of adopting the $\mathcal{L}$ are superior to $\mathcal{L}_{c}$ that is widely applied in well-annotated scenarios and $\mathcal{L}'$. This proves the presence of a considerable amount of noisy correspondence in PTM and the $\mathcal{L}$ adopted by RNCL contributes to the enhanced robustness of our RoMa. 3) DAP without any one of attention and positional embedding will decrease matching performance, demonstrating that each component in DAP contributes to the comprehensive perception of features.

\begin{figure*}
    \centering
    \includegraphics[width=1\linewidth]{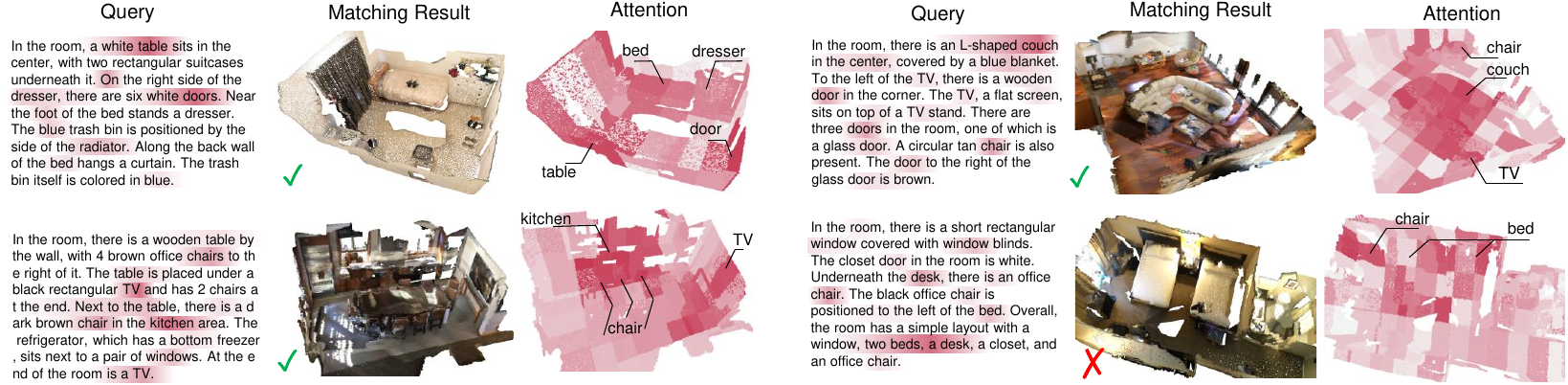}
     \caption{Some matched instances of PTM on SceneDepict-3D2T. For each text query, the top-1 ranked point cloud are displayed. The correctly matched point clouds are marked with a green tick, otherwise the red cross. In addition, we visualize the text after applying attention and present a comparison between the original point clouds and the point cloud after applying attention. \textcolor[HTML]{F43446}{\textbf{Redder}} words and patches indicate higher attention weights.}
    \label{vis_1}
    \vspace{-0.3cm}
\end{figure*}

\begin{figure}[h]
  \centering
  \subfloat[width=0.37\textwidth][Point cloud$\to$Text]{
		\includegraphics[height=2.75cm]{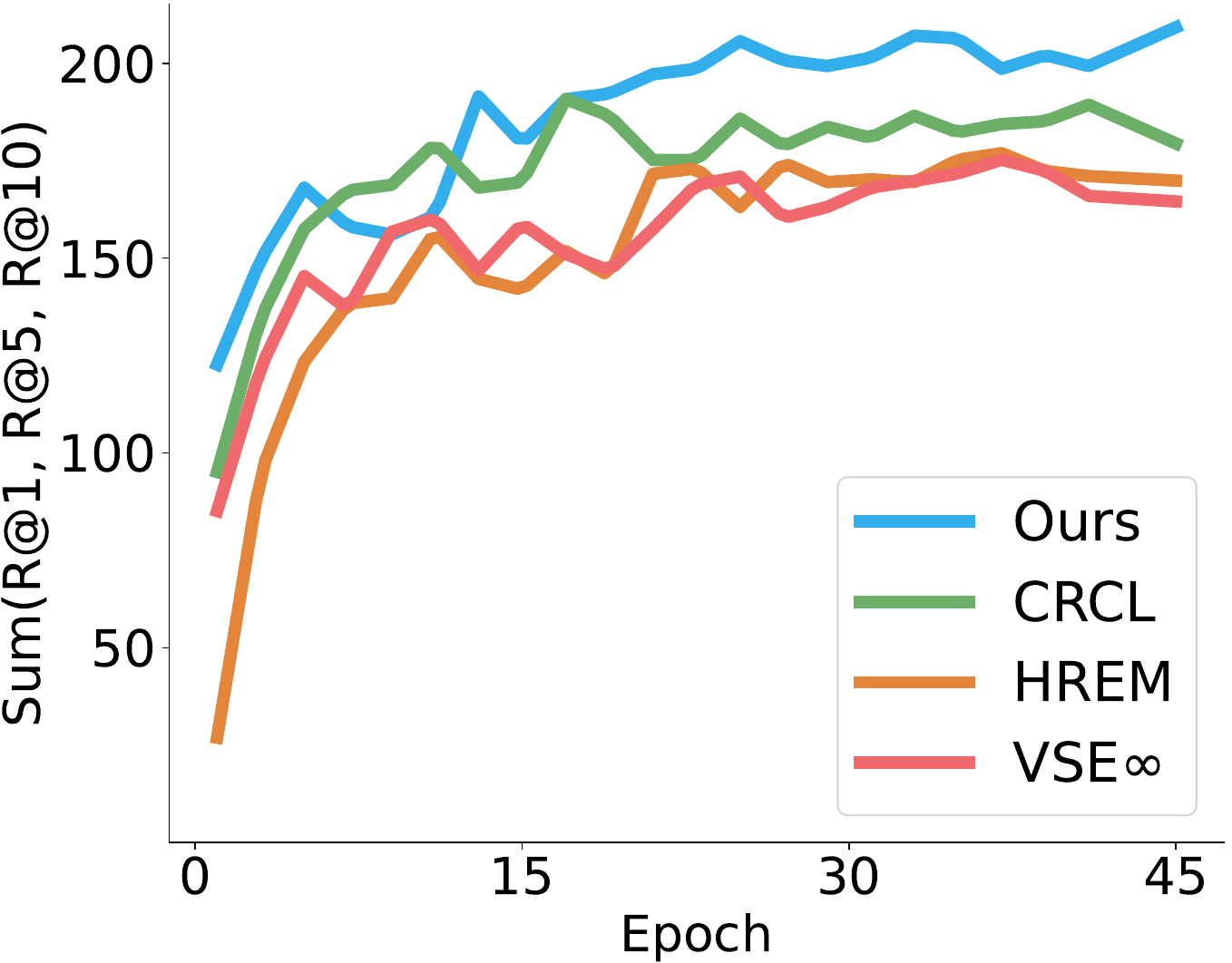}
		\label{FX1}
		}
  \hspace{0.2cm}
  \subfloat[width=0.37\textwidth][Text$\to$Point cloud]{
		\includegraphics[height=2.75cm]{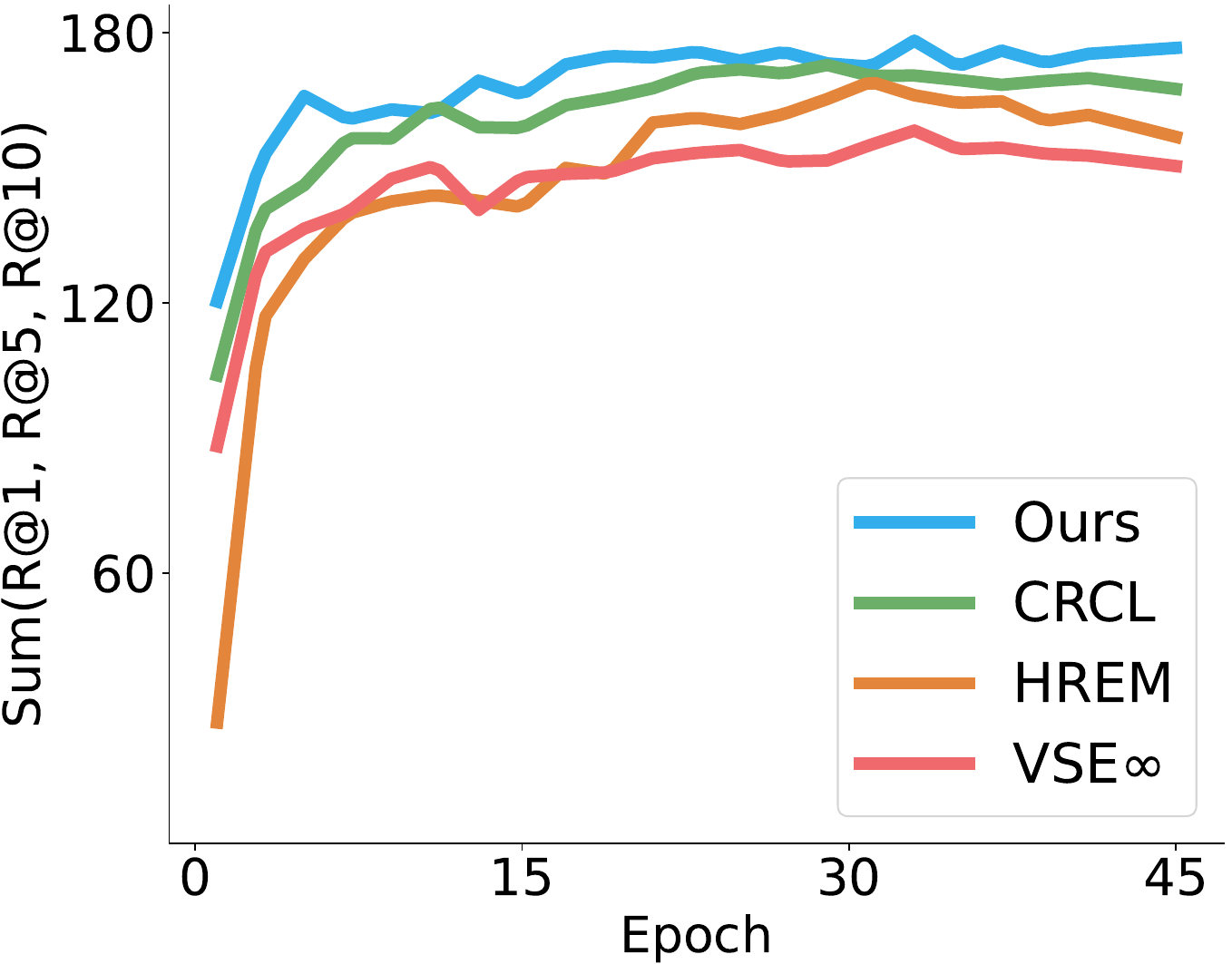}
		\label{FX2}
		}
  \caption{The performance of VSE$\infty$, HREM, CRCL, and our RoMa on the SceneDepict-3D2T dataset.}
  \label{vis_2}
  \vspace{-0.2cm}
\end{figure}

\subsection{Parameter Analysis}
To investigate the sensitivity of our RoMa to various parameters, we plot the sum of R@1, R@5, and R@10 of PTM against different hyper-parameters (i.e., $\alpha$, and $\tau$) on SceneDepict-3D2T as shown in \Cref{par}. The results indicate that our method achieves superior matching performance across a range of $\alpha$ and $\tau$ values. Notably, when $\alpha$ is set too low, the threshold for distinguishing between clean and noisy negative pairs becomes excessively small. This leads to a significant number of negative pairs being misclassified as noise and subjected to reverse optimization, resulting in remarkable performance degradation. Conversely, if $\alpha$ is too high, the RNCL struggles to differentiate potential noisy negative pairs, causing error accumulation and degraded performance.


\subsection{Visualization Analysis}
To provide a comprehensive insight into the effectiveness exhibited by our RoMa, we conduct visualization experiments in PTM. Firstly, to shed light on the reasons behind the superior performance of our RoMa, we illustrate a small handful of matching results and token-level attention visualization throughout the point clouds and texts on SceneDepict-3D2T dataset, as shown in~\Cref{vis_1}. Additionally, we present a performance comparison among our RoMa and the VSE$\infty$~\cite{chen2021learning}, HREM~\cite{Fu_2023_CVPR}, and CRCL~\cite{qin2024cross} throughout the training process, as shown in~\Cref{vis_2}. From the results, we could draw the following observations: 1) Our RoMa can achieve correct retrieved results in PTM. Even the mismatched pair still exhibits a strong cross-modal semantic correlation. This is attributed to our DAP, which actually focuses on useful and discriminative patches and words. 2) Throughout the whole learning process, it is evident that non-robust baselines (i.e., VSE$\infty$ and HREM) involve performance degradation in the later training stage, impacted by the noisy correspondence. In contrast, our RoMa mitigates the negative impact comprehensively, achieving superior and robust performance.

\section{Conclusion}
In this paper, we introduce a novel yet challenging task, named PointCloud-Text Matching (PTM). To facilitate the research on this promising task, we construct a benchmark dataset, namely SceneDepict-3D2T. We also propose a robust baseline, named \textbf{Ro}bust PointCloud-Text \textbf{Ma}tching method (RoMa), which consists of two novel modules: Dual Attention Perception module (DAP) and Robust Negative Contrastive Learning module (RNCL). Specifically, DAP leverages dual attention mechanisms to capture local and global features of point clouds and texts. In addition, RNCL is employed to handle noisy correspondence by distinguishing and endowing clean and noisy negative pairs with correct optimization directions. We conducted extensive experiments compared to 14 state-of-the-art methods on four datasets, demonstrating the superiority of our RoMa in the challenging PTM task.

\bibliography{main}{}
\bibliographystyle{ieeetr}

\end{document}